%% file: IJCAI2018_draft.tex
\documentclass{article}
\usepackage{ijcai18}
\usepackage[utf8]{inputenc} 
\usepackage{hyperref}       

\RequirePackage{amsmath,amssymb}
\usepackage{booktabs}       
\usepackage{amsfonts}       
\usepackage{nicefrac}       
\usepackage{microtype}      
\usepackage{times}
\usepackage{latexsym}
\usepackage{color}
\usepackage{amsmath}
\usepackage{amsfonts}
\usepackage{multirow}
\usepackage{times}
\usepackage{xcolor}
\usepackage{soul}
\usepackage[utf8]{inputenc}
\usepackage[small]{caption}
\usepackage{mathrsfs}
\usepackage{comment}
\usepackage{graphicx}
\usepackage{subcaption}
\usepackage{url}            

\begin{document}
\title{Improving Distantly Supervised Relation Extraction using Word and Entity Based Attention}
\author{
Sharmistha Jat$^1$ \thanks{Equal Contribution to this work}, 
Siddhesh Khandelwal$^2$ \footnotemark[1] \thanks{This research was conducted during the author’s research assistantship at Indian Institute of Science}, 
Partha Talukdar$^1$, 
\\ 
$^1$ Indian Institute of Science, Bangalore \\
$^2$ University of British Columbia\\
sharmisthaj@iisc.ac.in,
skhandel@cs.ubc.ca,
ppt@iisc.ac.in
}
\setlength\titlebox{2.5in}
\maketitle

\input{defs.tex}

\begin{abstract}
\input{abstract}
\end{abstract}

%
%

\input{introduction}

\input{model}

\input{dataset}

\input{experiments}

\input{related_work}
\input{conclusion}

\bibliographystyle{named}
\bibliography{IJCAI2018}

\end{document}

%% file: defs.tex
\newcommand{\refalg}[1]{Algorithm~\ref{#1}}
\newcommand{\refeqn}[1]{Equation~\ref{#1}}
\newcommand{\reffig}[1]{Figure~\ref{#1}}
\newcommand{\reftbl}[1]{Table~\ref{#1}}
\newcommand{\refsec}[1]{Section~\ref{#1}}
\newcommand{\method}[1]{\mbox{\textsc{#1}}}
\newcommand{\reminder}[1]{}

\newcommand{\systemwa}{BGWA}
\newcommand{\systemwafull}{Bi-GRU Word Attention}

\newcommand{\systemea}{EA}
\newcommand{\systemeafull}{Entity Attention}

\newcommand{\newdataset}{Google Distant Supervision}
\newcommand{\newdatasetshort}{GDS}

%% file: abstract.tex
Relation extraction is the problem of classifying the relationship between two entities in a given sentence. Distant Supervision (DS) is a popular technique for developing relation extractors starting with limited supervision. We note that most of the sentences in the distant supervision relation extraction setting are very long and may benefit from word attention for better sentence representation. Our contributions in this paper are threefold. Firstly, we propose two novel word attention models for distantly-supervised relation extraction: (1) a Bi-directional Gated Recurrent Unit (Bi-GRU) based word attention model (\systemwa{}), (2) an entity-centric attention model (\systemea{}), and (3) a combination model which combines multiple complementary models using weighted voting method for improved relation extraction. Secondly, we introduce \newdatasetshort{}, a new distant supervision dataset for relation extraction. \newdatasetshort{} removes test data noise present in all previous distant-supervision benchmark datasets, making credible automatic evaluation possible. Thirdly, through extensive experiments on multiple real-world datasets, we demonstrate the effectiveness of the proposed methods.

%% file: introduction.tex
\section{Introduction} 
Classifying the semantic relationship between two entities in a sentence is termed as Relation Extraction (RE). RE from unstructured text is an important step in various Natural Language Understanding tasks, such as knowledge base construction, question-answering etc. Supervised methods have been successful on the relation extraction task \cite{bunescu2005shortest,zeng:2014}. However, the extensive training data necessary for supervised learning is expensive to obtain and therefore restrictive in a Web-scale relation extraction task.

 To overcome this challenge, \cite{mintz:2009} proposed a Distant Supervision (DS) method for relation extraction to help automatically generate new training data by taking an intersection between a text corpus and knowledge base. The distant supervision assumption states that for a pair of entities participating in a relation, any sentence mentioning that entity pair in the text corpora is a positive example for the relation fact. This assumption outputs evidence from multiple sentences for multiple relation labels between an entity-pair. Therefore the problem of relation extraction in distantly supervised datasets is posed as a Multi-instance Multi-label (MIML) problem \cite{surdeanu:2012}, as shown in \reffig{fig:miml}. However, the DS assumption is too strong and may introduce noise such as false negative samples due to missing facts in the knowledge base. In this paper, we propose relation extraction models and a new dataset to improve RE. We define `instance' as a sentence containing an entity-pair, and `instance set' as a set of sentences containing the same entity-pair. 

It was observed by \cite{zeng:2015} that 50\% of the sentences in the Riedel2010 Distant Supervision dataset  \cite{riedel:2010}, a popular DS benchmark dataset, had 40 or more words in them. We note that not all the words in these long sentences contribute towards expressing the given relation. In this work, we formulate various word attention mechanisms to help the relation extraction model focus on the right context in a given sentence. 

The MIML assumption states that in an instance set corresponding to an entity pair, at least one sentence in that set should express the true relation assigned to the set. However, we observe that this is not always true in currently available benchmark datasets for RE in the distantly supervised setting. In particular, current datasets have noise in the \emph{test} set, for example, a fact may be labelled false if it is missing in the knowledge base, leading to a false negative label in train and test set. Noise in test set impedes the right comparison of models and may favor overfitted models. To address this challenge, we build the \newdataset{} (\newdatasetshort{}) dataset, a new dataset for distantly-supervised relation extraction. \newdatasetshort{} is seeded from the Google relation extraction corpus \cite{g:dataset}. This new dataset addresses an important shortcoming in distant supervision evaluation and makes an automatic evaluation in this setting more reliable. 

In summary, our contributions are: (a) we introduce the \newdataset{} (\newdatasetshort{}) dataset, a new dataset for distantly-supervised relation extraction; (b) we propose two novel word attention based models for distant supervision, viz., \systemwa{}, a BiGRU-based word attention model, and \systemea{}, an entity-centric attention model; and (c) we show efficacy of combining new and existing relation extraction models using a weighted ensemble model.

%% file: model.tex
\section{Proposed Methods}
\label{sec:proposed_methods}


In this section, we present our attention-based models for distantly supervised relation extraction. We first describe problem background and Piecewise Convolution Neural Network (PCNN), a previous-state-of-the-art model. We then introduce our Entity attention (EA) and Bi-GRU based word attention (BGWA) models. The last subsection describes a simple ensemble approach to combine predictions of various models for robust relation extraction.  


\subsection{Background}
\label{sec:background}

\begin{figure}[t]
  \centering
  \setlength{\textfloatsep}{0.42cm}
  \includegraphics[scale=0.26]{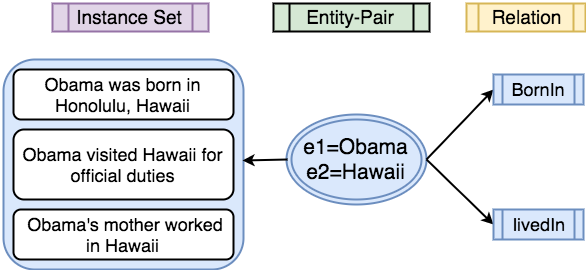}
  \caption{\label{fig:miml}Multi-instance Multi-label (MIML) learning sample instance for Relation Extraction in Distant Supervision.}
\end{figure}

\begin{figure}[t]
  \centering
  \setlength{\textfloatsep}{0.30cm}
  \includegraphics[scale=0.24]{pcnn}
  \caption{Piecewise max-pooling in the PCNN model proposed by \protect\cite{zeng:2015}. The two entities are underlined. (\refsec{sec:background})}
  \label{fig:pcnn}
\end{figure}
\begin{figure}[t]
   \centering
   \setlength{\textfloatsep}{0.30cm}
   \includegraphics[width=5.8cm,height=4.2cm]{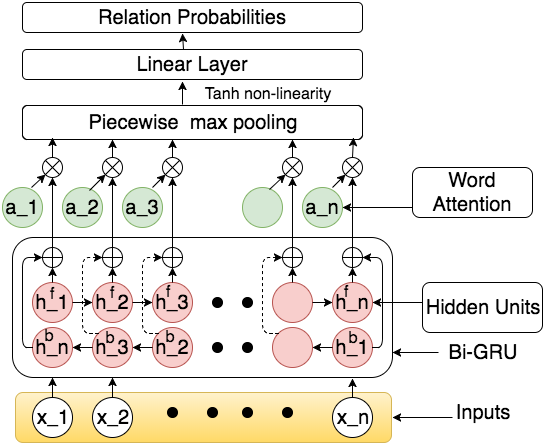}
   \caption{\label{gru_att}Bi-GRU word attention (BGWA) model  (\refsec{sec:wordatt})}
\end{figure}%
~~~~\textbf{Relation Extraction}: A relation is defined as a semantic property between a set of entities $\{e_k\}$. In our task, we consider binary relations where $k \in [1,2]$, such as Born\_In(Barack Obama, Hawaii). Given a set of sentences $S = \{s_i\}; i \in [1 \ldots N]$, where each sentence $s_i$ contains both the entities, the task of relation extraction with distantly supervised dataset is to learn a function $F_r$:
\[
  F_r(S,(e_1,e_2)) = \left.
  \begin{cases}
    1\ \text{if relation r is true for pair} (e_1, e_2)\\
    0\ \text{Otherwise} \\
    
  \end{cases}
  \right\} 
\]
        
\textbf{PCNN}: \cite{zeng:2015} proposed the Piecewise Convolution Neural Network (PCNN), a successful model for distantly supervised relation extraction. The Success of the relation extraction task depends on extracting the right structural features from the sentence containing the entity-pair. Neural networks, such as Convolutional Neural Networks (CNNs), have been proposed to alleviate the need to manually design features for a given task \cite{zeng:2014}. As the output of  CNNs is dependent on the number of tokens in the sentence, max pooling operation is often applied to remove this dependence. However, the use of a single max-pool misses out on some of these structural features useful for relation extraction task. PCNN model divides a sentence $s_i$'s convolution filter output $c_i$ containing two entities into three parts $c_{i1},c_{i2},c_{i3}$-- sentence context to the left of first entity, between the two entities, and to right of the second entity respectively-- and performs max-pooling on each of the three parts, shown in \reffig{fig:pcnn}. Thereby, leveraging the entity location information to retain the structural features of a sentence after the max-pooling operation.
\[
    pc_{ij} = max(c_{ij}); 1 \le i \le N, 1 \le j \le 3 
\]
The output of this operation is the concatenation of $\{ pc_{i1},pc_{i2},pc_{i3} \}$ yielding a fixed size output. The fixed size output is processed through a tanh non-linearity followed by a linear layer to produce relation probabilities.

\subsection{Bi-GRU based Word Attention Model (\systemwa{})}
\label{sec:wordatt}

Consider the sentence expressing \textit{bornIn(Person, City)} relation between the entity pair \textit{(Obama, Honolulu)}.

\begin{center}
\textit{Former President Barack \textbf{Obama} was born in the city of \textbf{Honolulu}, capital of the U.S. state of Hawaii}
\end{center}

 In the sentence, the phrase \textit{``was born in"} helps in identifying the correct relation in the sentence. It is conceivable that identifying such key phrases or words will be helpful in improving relation extraction performance. Motivated by this, our first proposed model, Bidirectional Gated Recurrent Unit (Bi-GRU) based Word Attention Model (\systemwa{}) uses an attention mechanism over words to identify such key phrases. To the best of our knowledge, there has been no prior work on using word attention in the distant supervision setting.

The \systemwa{} model is shown in \reffig{gru_att}. It uses Bi-GRU to encode sentence context. GRU \cite{cho2014properties} is a variant of Recurrent Neural Network (RNN) which was designed to capture long-range dependencies in words. A Bi-GRU runs in both forward and backward direction in a sentence to capture both sides of a word context. 

Only a few words in a sentence are relevant for determining the relation expressed. This degree of relevance of a word is calculated as an attention value in our model. An instance set $S_q$ consists of set of sentences, $[{s_i}; i \in [1 \ldots N]]$. Each word in sentence $s_i$ is represented using a pre-trained embedding $x_{ij} \in \mathbb{R}^{d \times 1}, j \in [1 \ldots M]$. Passing $s_i$ through GRU generates forward ($h^{f}_{ij}$) and backward ($h^{b}_{ij}$) hidden representations for each word. Concatenation $[h^{f}_{ij}, h^{b}_{ij}] \in \mathbb{R}^{g\times 1}$ is used as representation for a word.
\begin{align*}
  w_{ij} = [h^{f}_{ij}, h^{b}_{ij}];~~~~u_{ij} = w_{ij} \times A \times r ; \\
  a_{ij} = \frac{\exp(u_{ij})}{\displaystyle\sum_{l=1}^{M}{\exp(u_{il})}};~~~~\hat{w_{ij}} = a_{ij} \times w_{ij}
\end{align*}

We define $u_{ij}$, the degree of relevance of the $j^{th}$ word in $i^{th}$ sentence of the instance set, \noindent where $A \in R^{g \times g}$ is a square matrix and $r \in \mathbb{R}^{g \times 1}$ is a relation query vector. Bilinear operator $A$ determines the relevance of a word for a relation vector. Both $A$ and $r$ are learned parameters. Attention value $a_{ij}$ is calculated by taking softmax over $\{u_{ij}\}$ values. 
    
Despite the widespread use of weighted sum to obtain sentence context embeddings $s_{wa}$ in attention-based settings, similar to PCNN model \refsec{sec:background}, we apply the piecewise max pooling on $\hat{w_{ij}}$ before, between, and after the entity pair.  Final piecewise max-pooled sentence embedding $s_{wa} \in \mathbb{R}^{1 \times 3g}$ of the sentence is processed through a $tanh$ non-linearity and linear layer to yield probabilities for each relation.

\begin{figure}[t]
  \centering
\setlength{\textfloatsep}{0.30cm}
\includegraphics[scale=0.24]{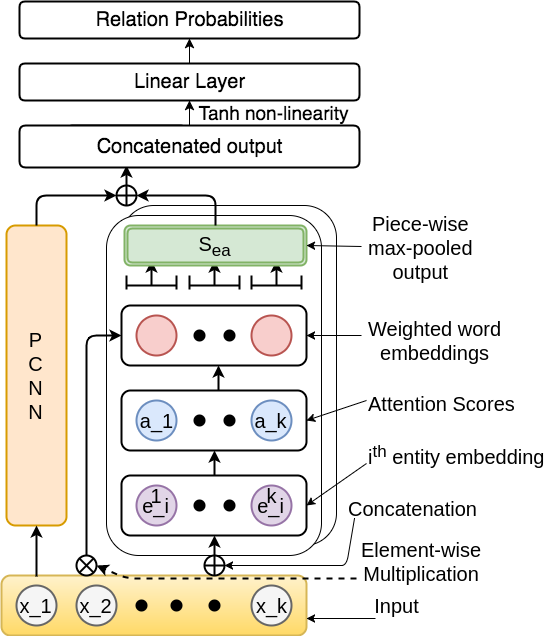}
  \caption{\label{EAatt}Entity Attention (EA) Model (\refsec{sec:entatt})}
\end{figure}

\subsection{Entity Attention (\systemea{}) Model}
\label{sec:entatt}

Let us once again consider the example sentence from \refsec{sec:wordatt} involving entity pair \textit{(Obama, Honolulu)}. In the sentence, for entity \textit{Obama}, the word \textit{President} helps in identifying that the entity is a person. This extra information helps in narrowing down the relation possibilities by looking only at the relations that occur between a person and a city. \cite{byshen:2016} proposed an entity attention model for \emph{supervised} relation extraction with a single sentence as input to the model. We modify and adapt their model for the \emph{distant supervision} setting and propose \systemeafull{} (\systemea{}) which works with a bag of sentences. For a given bag of sentences, learning is done using the setting proposed by \cite{zeng:2015}, wherein the sentence with the highest probability of expressing a relation in a bag is selected to train the model in each iteration.

The \systemea{} model has two components: 1) PCNN layer, and 2) Entity Attention Layer, as shown in  \reffig{EAatt}. Consider an instance set $S_q$ with set of sentences, $[{s_i}; i \in [1 \ldots N]]$ and an entity-pair $e_{qk}, k \in [1,2]$. A sentence $s_i$ has $M$ words $[x_{ij}; j \in [1 \ldots M]]$, where each $x_{ij} \in \mathbb{R}^{1 \times d}$ is a word embedding and $\{e_{q1}^{emb}, e_{q2}^{emb}\}$ are the embeddings for the two entities. The PCNN layer is applied on the words in the sentence \cite{zeng:2015}. The entity-specific attention $u_{i,j,qk}$ for $j^{th}$ word with respect to $k^{th}$ entity is calculated as follows:
\[
        u_{i,j,qk} = [x_{ij}, e_{qk}^{emb}] \times A_{k} \times  r_{k},
\]
Here, $[x_{ij}, e_{qk}^{emb}]$ is the concatenation of a word and the entity embedding. $A_{k}$, $r_{k}$ are learned parameters. Bilinear operator $A_{k}$ determines the relevance of concatenated word \& entity embedding for a relation vector $r_{k}$. Intuitively, attention should choose words which are related to the entity for a given relation. The $u_{i,j,qk}$ are normalized using a softmax function to generate $a_{i,j,qk}$, the attention scores for a given word. Similar to the PCNN model in \refsec{sec:background}, the attention weighted word embeddings are pooled using piecewise pooling method to generate $s_{ea} \in \mathbb{R}^{1 \times 3g}$ dimensional sentence embeddings. The output from the PCNN layer and the entity attention layers are concatenated and then passed through a linear layer to obtain probabilities for each relation.

The entity attention model (\systemea{}) we propose is adapted to the distantly supervised setting by using two important variations from the original \cite{byshen:2016} model (a) The \systemea{} processes a set of sentences. It uses PCNN \cite{zeng:2015} assumption to select the sentence with highest probability of any relation. The selected sentence is used to estimate the relation probabilities for an entity-pair and for back-propagation of the error for the bag-of-sentences. (b) \systemea{} uses PCNN instead of CNN to preserve structural features in a sentence. We found the two variations to be crucial for the model to work in the distant supervision setting. 



\subsection{Bring it all together: Ensemble Model}
\label{sec:ensemble}

\begin{figure}[t]
        \centering
        \setlength{\textfloatsep}{0.30cm}
        \includegraphics[width=3.1cm,height=3.3cm]{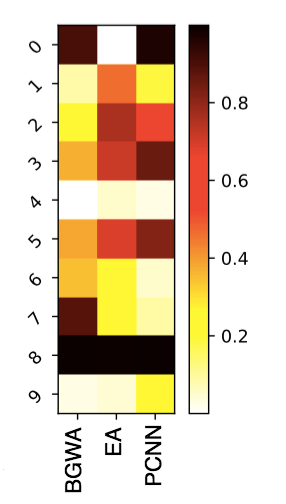}
        

\caption{ Confidence scores (indicated by color intensity, darker is better) of models on true labels of 10 randomly sampled instance sets from \newdataset{} dataset. Rows represent the instance sets and columns represent the model used for prediction. The heatmap shows complementarity of these models in selecting the right relation. Motivated by this evidence, the proposed Ensemble method combines the three models, viz., Word Attention (BGWA), Entity Attention (EA) and PCNN.}
\label{fig:heatmap}
\end{figure}

We note that the models discussed in previous sections, \systemwa{}, \systemea{} and PCNN, have complementary strengths. PCNN extracts high-level semantic features from sentences using CNN. Most effective features are then selected using a piecewise max-pooling layer. Entity-based attention (\refsec{sec:entatt}) helps in highlighting important relation words with respect to each of the entities present in the sentence, thus complimenting the PCNN-based features. Going beyond the entity-centric words, we observe that not all words in a sentence are equally important for relation extraction. The \systemwa{} model (\refsec{sec:wordatt}) addresses this aspect by selecting words relevant to a relation in a sentence. 

In \reffig{fig:heatmap}, we plot the confidence scores of various models on the true labels of 10 randomly selected instance sets from \newdataset{} dataset (described in \refsec{sec:dataset}). From this figure, we observe that the proposed methods are able to leverage signals from the entity and word attention models, even when the PCNN model is incorrect (light colored cell in the last column). This validates our assumption and motivates ensemble approach to efficiently combine these complementary models.

We combine the predictions of all the three models using a weighted voting ensemble. The weights of this model are learned using linear regression on development dataset. Assume $P_{i,<model>}$ is a vector containing probability scores for all relations with respect to $i^{th}$ example in developement data as given by a model. $P_{i,<model>} \in \mathbb{R}^{1 \times rl}$, where $rl$ is the number of relations.  
\[
P_{i,ENSEMBLE} = \alpha * P_{i,PCNN} + \beta * P_{i,\systemea{}} + \gamma * P_{i,\systemwa{}}
\]

Here, $\alpha, \beta, \gamma$ are parameters learned using linear regression \cite{scikit-learn}. More complicated regression methods (e.g., ridge regression) did not improve the results greatly. We also experimented with a jointly learned neural ensemble by concatenating the features of all models after pooling layer followed by a linear layer. In our experiments, weighted voting ensemble method gave better results than the jointly learned model.

%% file: dataset.tex
\section{\newdatasetshort{}: A New Dataset for Relation Extraction using Distant Supervision}
\label{sec:dataset}

 \begin{table*}[t]
 \centering
  \scriptsize
\begin{tabular}{|c|c|c|c|c|}
\hline

\textbf{S.No.}&\textbf{Entity 1} & \textbf{Entity 2} & \textbf{Test Set Label} & \textbf{Classified Relation}  \\ \hline

1. & Marlborough       & New Hampshire    & NA  & /location/location/contains \\ \hline
2. & Katie Couric      & CBS             & NA   & /business/person/company \\ \hline
\textbf{S.No.}&\textbf{Entity 1} & \textbf{Entity 2}  & \textbf{Test Set Label}  & \textbf{Instance Set}\\ \hline
3. &\begin{tabular}[c]{@{}c@{}} Gary \\Sheffield  \end{tabular}     & Florida     & \begin{tabular}[c]{@{}c@{}}/people/person/\\place\_lived \end{tabular}  & 
\begin{tabular}[c]{@{}c@{}}others who have already indicated they will wear no. 42 include ken griffey jr. of cincinnati,\\florida's dontrelle willis, carlos lee of houston, derrek lee of the cubs and detroit's gary\_sheffield .\end{tabular}\\ \hline
4. &\begin{tabular}[c]{@{}c@{}} Brian \\Dawkins  \end{tabular}     & Jacksonville     & \begin{tabular}[c]{@{}c@{}}/people/person/\\place\_of\_birth \end{tabular}  & 
\begin{tabular}[c]{@{}c@{}}according to glazer, philadelphia's brian dawkins and jacksonville's \\ donovin darius have trained at a mixed martial arts gym. \end{tabular}\\ \hline
\end{tabular}
\caption{\label{tab:noiseexamples}Examples of Noise in dataset. Sample 1,2 are incorrectly labelled with NA relation in the test set due to missing facts in KB. While, Sample 4 \& 5's single sentence in the instance set does not support the KB relation.}

\end{table*}

\begin{table}[t]
      \scriptsize
     \centering
    \begin{tabular}{|c|c|c|}
    \hline
    Relation - Class           & No. sentences   & No. entity-pair \\ \hline
    \textit{perGraduatedInstitution}     & 4456 & 2624  \\ \hline
    \textit{perHasDegree} & 2969 & 1434   \\ \hline
    \textit{perPlaceOfBirth}      & 3356        & 2159   \\ \hline
    \textit{perPlaceOfDeath}    & 3469     & 1948 \\ \hline
    NA      & 4574 & 2667 \\ \hline
    \end{tabular}
    \caption{\label{tab:dataset}Statistics of the new  \newdatasetshort{} dataset. Please see \refsec{sec:dataset} for more details.}
\end{table}

 \begin{figure}[t]
  \centering
  \setlength{\textfloatsep}{0.30cm}
  \includegraphics[width=5.1cm,height=2.7cm]{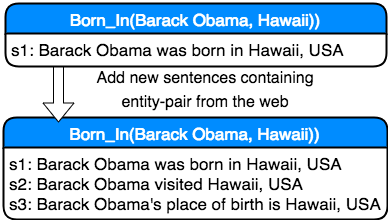}
  \caption{\label{fig:gids_creation}Example of instance set creation in GDS dataset.}
\end{figure}


Several benchmarks datasets for Relation Extraction (RE) using distant supervision (DS) exist \cite{riedel:2010,mintz:2009}. DS is used to create both train and test sets in all of these datasets, resulting in the introduction of noise. While training noise in distant supervision is expected, noise in the test data is troublesome as it may lead to incorrect evaluations. There are two kinds of noise added due to distant supervision assumption: (a) samples with incorrect labels due to missing Knowledge Base (KB) fact, and (b) samples with no instance supporting the KB fact. A few examples of such noise are listed in \reftbl{tab:noiseexamples}. Previous benchmark datasets in this area suffer from these drawbacks. 

In order to overcome these challenges, we develop \newdataset{} (\newdatasetshort{}), a new dataset for relation extraction using distant supervision. Statistics of the new dataset are summarized in \reftbl{tab:dataset}. To alleviate noise in DS setting, we make sure that labelled relation is correct and for each instance set in \newdatasetshort{}, there is at least one sentence in that set which expresses the relation assigned to that set. 

We start with the human-judged Google Relation Extraction corpus\footnote{https://research.googleblog.com/2013/04/50000-lessons-on-how-to-read-relation.html}. This corpus consists of 5 binary relations. We construct the \newdatasetshort{} dataset out of the relation extraction corpus using the following process. Let $D_{\mathrm{GRE}}$ be the Google RE corpus, $D_{\mathrm{GRE}} = \{(\mathbf{s}_i, e_{i1}, e_{i2}, r_i)\}$ , where the $i^{th}$ sentence $\mathbf{s}_i$ is annotated as expressing relation $r_i$ between the two entities $e_{i1}$ and $e_{i2}$ in the sentence. $r_i$ is one of the five relations mentioned in \reftbl{tab:dataset}. Now, for each $(\mathbf{s}_i, e_{i1}, e_{i2}, r_i) \in D_{\mathrm{GRE}}$, we perform the following:

\begin{itemize}
  \item Perform web search to retrieve documents containing the two entities $e_{i1}$ and $e_{i2}$.
  \item From such retrieved documents, select multiple text snippets containing the two entities. Each snippet is restricted to contain at most 500 words. Let $S_i = \{\mathbf{s}_q^{'}\}; q \in (1 \ldots M^{'})$ be the set of such snippets.
  \item Let $S_{i}^{'} = \{\{\mathbf{s}_i\} \cup S_i\}$. We now create a new instance set $B_i = \{(S_{i}^{'}, e_{i1}, e_{i2}, r_i)\}$. Here, $B_i$ is an instance set for distant supervision which consists of the set of instances (sentences or snippets) $S_{i}^{'}$, where the entities $e_{i1}$ and $e_{i2}$ are mentioned in each instance. The label $r_i$ is applied over the entire set $B_i$.
\end{itemize}

$D_{\mathrm{GDS}} = \{B_i\}$ is the new \newdatasetshort{} dataset. Here, each set $B_{i}$ is guaranteed to contain at least one sentence ($\mathbf{s}_i$) which expresses the relation $r_i$ assigned to that set. An example of the sentence set expansion is shown in \reffig{fig:gids_creation}.   We note that such guarantee was not available in previous DS benchmarks.

We divided this dataset into a train (60\%), development (10\%) and test (30\%) sets, such that there is no overlap among entity-pairs of these sets. Unlike currently available datasets, the availability of development dataset helps in performing model selection in a principled manner for relation extraction. 

In \cite{riedel:2010} and subsequent work, a manual evaluation was done by validating the top 1000 confident predictions. This manual evaluation was necessary due to the noise in the test data. \newdatasetshort{} although a small dataset in terms of the size as compared to Riedel2010 dataset, gets past such cumbersome manual evaluation and makes an automated evaluation in distantly-supervised relation extraction a reality.

%% file: experiments.tex
\section{Experiments and Results}
\label{sec:expts}

 \textbf{Datasets}: We validate effectiveness of the proposed models on two datasets summarized in \reftbl{tab:ds_dataset_summary}. {\bf Riedel2010} was created by aligning Freebase relations with the New York Times corpus \cite{riedel:2010,hoffmann:2011,surdeanu:2012,nre-lin} . We partitioned Riedel2010 train set into a new train (80\%) and development set (20\%). Development set is created to facilitate the learning of an ensemble model and for model selection. This resulting dataset is called {\bf Riedel2010-b}. Details of the new {\bf \newdatasetshort{}} dataset is described in \refsec{sec:dataset}. 

\textbf{Evaluation Metrics}: Following \cite{nre-lin}, we use held-out evaluation scheme. The performance of each model is evaluated on a test set using Precision-Recall (PR) curve. 

\textbf{Baselines}: We compare proposed models with (a) Piecewise Convolution Neural Network (PCNN) \cite{zeng:2015} and (b) Neural Relation Extraction with Selective Attention over Instances (NRE) \cite{nre-lin}. Both NRE and PCNN baseline outperform traditional baselines like MIML-RE and hence we use them as a representative state-of-the-art baseline to compare with proposed models.

\textbf{Model Parameters}: The parameters used for the various models are summarized in \reftbl{tab:paramters}. Word embeddings are initialized using the Word2Vec vectors from NYT dataset, similar to \cite{nre-lin}. Word Position feature embeddings (with respect to each entity) are randomly initialized and learned during training. Concatenation of the word embedding and position embedding results in a 60-dimensional ($d_w+(2*d_p)$) embedding $x_{ij}$ for each word.  We implemented PCNN model baseline following \cite{zeng:2015} and used author provided results and implementation for NRE baseline. The \systemea{} and \systemwa{} models were developed in PyTorch\footnote{\url{http://pytorch.org/}}. We use SGD algorithm with dropout \cite{srivastava2014dropout} for model learning. 
The experiments were run on GeForce GTX 1080 Ti using NVIDIA-CUDA. Model selection for all algorithms was done based on the AUC (Area Under the Curve) metric for the precision-recall curve for development dataset. 

\begin{table}[t]
      \scriptsize
      \centering
        \begin{tabular}{|c|c|c|c|}
          \hline
          Dataset           & \# relation & \# sentences   & \# entity-pair \\ \hline
          \multicolumn{4}{|l|}{\textbf{Reidel2010-b} Dataset with development set} \\ \hline
          Train & 53 &  455,771 &  233,064 \\ \hline
          Dev & 53 &  114,317 & 58,635  \\ \hline
          Test & 53 & 172,448 & 96,678 \\ \hline
          \multicolumn{4}{|l|}{\textbf{\newdatasetshort{}}  Dataset} \\ \hline
          Train & 5 & 11297 & 6498 \\ \hline          
          Dev & 5 & 1864 & 1082\\ \hline
          Test & 5 & 5663 & 3247 \\ \hline
        \end{tabular}
        \caption{\label{tab:ds_dataset_summary}Statistics of various datasets used in the paper.}
\end{table}
\begin{table}[t]
      \scriptsize
      \centering
        \begin{tabular}{|c|c|}
          \hline        
          Word Embedding Dimension & 50 \\ \hline
          Word Postion Emb Dimension & 5 \\ \hline
          Batch Size & 50 \\ \hline
          SGD Learning Rate & 0.1 \\ \hline
          Dropout Rate & 0.5 \\ \hline
        \end{tabular}
        \caption{\label{tab:paramters}Parameter settings}
\end{table}


\begin{figure}[t]
\centering
  \setlength{\textfloatsep}{0.30cm}
  \includegraphics[width=7.4cm,height=3.7cm]{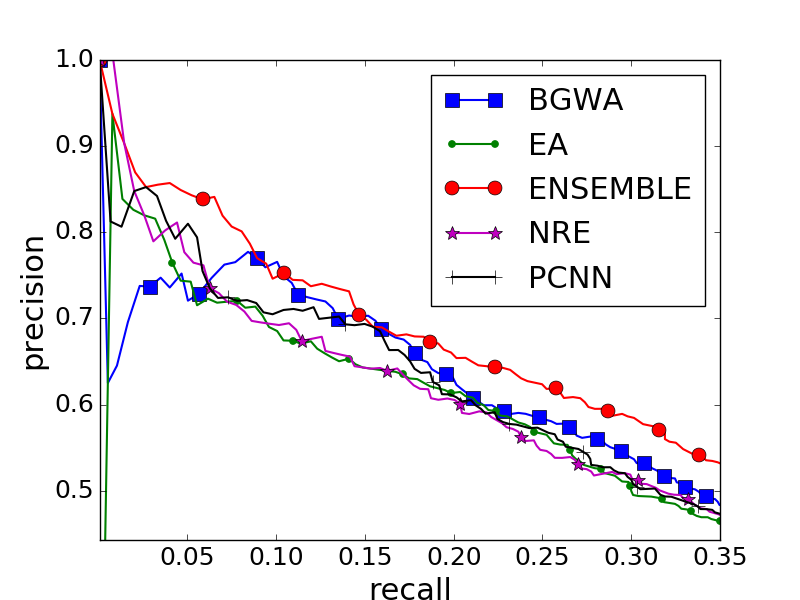}
  \caption{\label{fig:nre_dataset_results}Precision-recall curves of proposed models against traditional state-of-the-art methods on the Riedel2010-b dataset. 
}
 \end{figure}
 
\begin{figure}[t]
\centering
  \setlength{\textfloatsep}{0.30cm}
  \includegraphics[width=7.4cm,height=3.7cm]{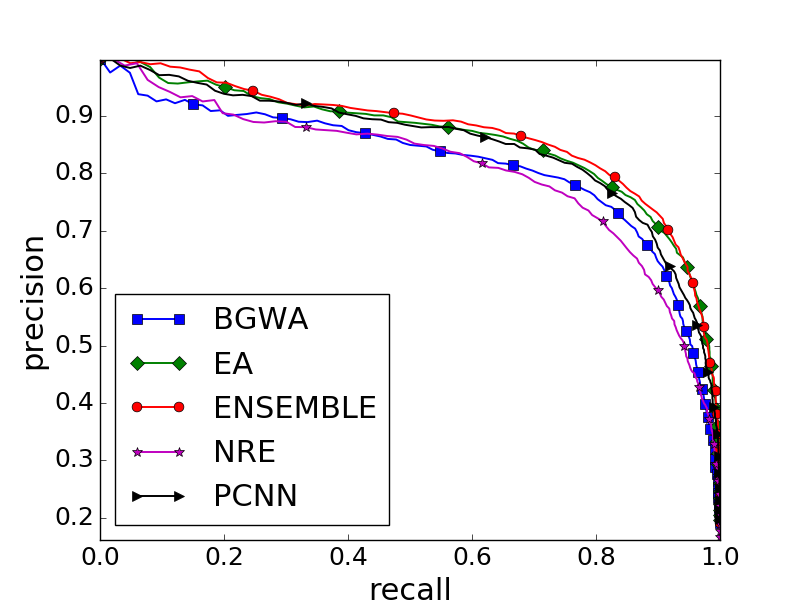}
  \caption{\label{fig:new_dataset_results}Precision-recall curves of various models on the \newdatasetshort{} dataset. Please see \refsec{sec:exp_results} for details.}
\end{figure}

\begin{figure}[t]
\centering
  \setlength{\textfloatsep}{0.30cm}
  \includegraphics[width=5.9cm,height=2.7cm]{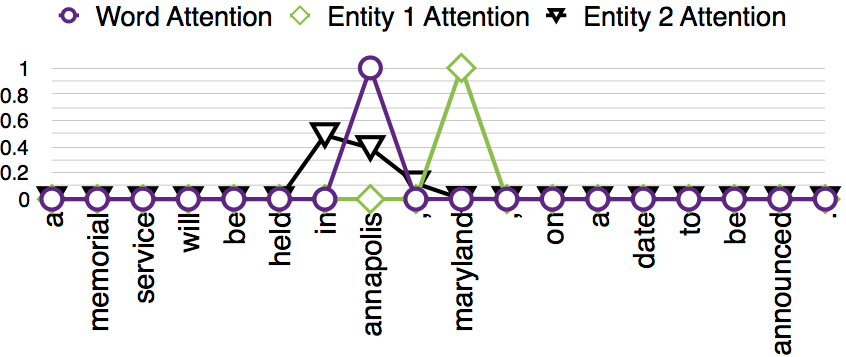}
  \caption{\label{fig:attviz}BGWA (word attention) and EA (entity attention) values for an example sentence between entity pair (\textit{maryland},  \textit{annapolis}) and relation \textit{location\_in}. X-axis shows the sentence words and y-axis shows the attention scores. Please see \refsec{sec:exp_results} for discussion.}
\end{figure}

\subsection{Results}
\label{sec:exp_results}

\textbf{Performance Comparison}: \reffig{fig:nre_dataset_results} and \reffig{fig:new_dataset_results} show the precision-recall curve for baseline and proposed algorithms on two datasets, Riedel2010-b (with development set) 
and the \newdatasetshort{} dataset. Please note that the NRE model's PR-curve in \reffig{fig:nre_dataset_results} is taken from author published results which used  combined train+dev set for training. 
This gives the NRE model 
an advantage over all other models in 
\reffig{fig:nre_dataset_results} as all of them are trained using only the train part. For the Riedel2010-b dataset, we plot the PR-curve with a maximum recall of 0.35, as the precision is too low beyond 0.35 recall.

From \reffig{fig:nre_dataset_results} and \reffig{fig:new_dataset_results}, we observe that the proposed models -- \systemwa{} and \systemea{} --  achieve higher or competitive precision over the entire recall range compared to the state-of-the-art NRE and PCNN models. PCNN model outperforms NRE model in both datasets. ENSEMBLE, a combination of proposed models \systemwa{}, \systemea{} and PCNN in a weighted ensemble, helps in improving precision further. It achieves a significant precision gain of over 2-3\% over various recall ranges for the Riedel2010-b dataset with 53 relations. This indicates that clues from combined model help results.

We observe that the \systemwa{} model performs well on the Riedel2010-b dataset, but the trend is reversed in performance for \newdatasetshort{} dataset where \systemea{} performs better. These two datasets have varied properties, (a) Riedel2010-b has 53 relations as opposed to 5 in the \newdatasetshort{} dataset, (b) \newdatasetshort{} has no label noise in the test as compared to the Riedel2010-b dataset. The performance difference between the \systemwa{} and \systemea{} model shows that the errors by both the models are not correlated and are complimentary as shown in \reffig{fig:heatmap}. This empirical validation encourages ensemble of these methods. We observe that the ENSEMBLE model performs consistently well across all recall ranges in both the datasets, validating our assumption. 

\textbf{Visualizing Attention}: We visualize the attention values of our models in \reffig{fig:attviz}. It can be observed that the Entity 2 Attention for the `\textit{location\_in}' relation rightly focuses on words indicating place information like `\textit{in}', `\textit{,}' and `\textit{annopolis}'. We note that entity attention is relation specific. In this case, Entity 1 Attention rightly focuses on the second entity,  `\textit{maryland}' (location name), for selecting relation `\textit{location\_in}'.  The word attention value is calculated using Bi-GRU hidden representation embeddings. Bi-GRU representation at a given time point $t$ in a sequence is a summary of all the timepoints correlated to $t$, in the sequence. A high attention value for the hidden layers after processing the word `\textit{annapolis}' indicates that the sentence has rich context around the first entity to indicate \textit{location\_in} relation. In conclusion, the attention models rightly choose the relevant words in context and help in improving relation extraction performance.

%% file: related_work.tex
\section{Related Work}

 Relation extraction in distantly supervised datasets is posed in a Multi-instance Multi-label (MIML) setting \cite{surdeanu:2012}. A large proportion of the subsequent work in this field has aimed to relax the strong assumptions that the original DS model made. \cite{riedel:2010} introduced the \textit{expressed-at-least-once} assumption in a factor graph model as an aggregating mechanism over mention level predictions. Work by \cite{hoffmann:2011,surdeanu:2012,ritter:2013} are crucial increments to \cite{riedel:2010}. 

 In past few years, Deep learning models \cite{Bengio:2009:LDA:1658423.1658424} have reduced the dependence of algorithms on manually designed features. \cite{zeng:2014} introduced the use of a CNN based model for relation extraction. \cite{zeng:2015} proposed a Piecewise Convolutional Neural Network (PCNN) model to preserve the structural features of a sentence using piecewise max-pooling approach, improving the precision-recall curve significantly. However, PCNN method used only one sentence in the instance-set to predict the relation label and for backpropagation. \cite{nre-lin} improves upon PCNN results by introducing an attention mechanism to select a set of sentences from instance set for relation label prediction.

  \cite{zheng:2016} aimed to leverage \textit{inter-sentence} information for relation extraction in a ranking model. The hypothesis explored is that for a particular entity-pair, each mention alone may not be expressive enough of the relation in question, but information from several mentions may be required to decisively make a prediction. Recently, work by \cite{DBLP:journals/corr/YeCL16} exploit the connections between relation (class ties) to improve relation extraction performance.

A few papers propose the addition of background knowledge to reduce noise in training data. \cite{weston:2013} proposes a joint-embedding model for text and KB entities where the known part of the KB is utilized as part of the supervision signal. \cite{han:2016} use indirect supervision like consistency between relation labels, consistency between relations and arguments, and consistency between neighbour instances using Markov logic networks. \cite{candis} uses inter-instance-set couplings for relation extraction in multi-task setup to improve performance.

 Attention models learn the importance of a feature in the supervised task through back-propogation. Attention mechanisms in neural networks have been successfully applied to a variety of problems, like machine translation \cite{bahdanau2014neural}, image captioning \cite{xu+al-2015-icml}, supervised relation extraction \cite{byshen:2016}, distantly-supervised relation extraction \cite{zheng:2016} etc. 

In our work, we focus on selecting the right words in a sentence using the word and entity-based attention mechanism. 

%% file: conclusion.tex
\section{Conclusion}

Distant Supervision (DS) has emerged as a promising approach to bootstrap relation extractors with limited supervision. In this paper, we present three novel models for distantly-supervised relation extraction: (1) a Bi-GRU based word attention model (\systemwa{}), (2) an entity-centric attention model (\systemea{}), and (3) and a weighted voting ensemble model, which combines multiple complementary models for improved relation extraction. We introduce \newdatasetshort{}, a new distant supervision dataset for relation extraction. \newdatasetshort{} removes test data noise present in all previous distant supervision benchmark datasets, making credible automatic evaluation possible. Combining proposed methods with attention-based sentence selection methods is left as future work. We plan to make our code and datasets publicly available to foster reproducible research.

%% file: IJCAI2018_draft.bbl
\begin{thebibliography}{}

\bibitem[\protect\citeauthoryear{Bahdanau \bgroup \em et al.\egroup
  }{2014}]{bahdanau2014neural}
Dzmitry Bahdanau, Kyunghyun Cho, and Yoshua Bengio.
\newblock Neural machine translation by jointly learning to align and
  translate.
\newblock {\em arXiv preprint arXiv:1409.0473}, 2014.

\bibitem[\protect\citeauthoryear{Bengio}{2009}]{Bengio:2009:LDA:1658423.1658424}
Yoshua Bengio.
\newblock Learning deep architectures for ai.
\newblock {\em Found. Trends Mach. Learn.}, 2(1):1--127, January 2009.

\bibitem[\protect\citeauthoryear{Bunescu and
  Mooney}{2005}]{bunescu2005shortest}
Razvan~C Bunescu and Raymond~J Mooney.
\newblock A shortest path dependency kernel for relation extraction.
\newblock In {\em Proceedings of the conference on human language technology
  and empirical methods in natural language processing}, pages 724--731.
  Association for Computational Linguistics, 2005.

\bibitem[\protect\citeauthoryear{Cho \bgroup \em et al.\egroup
  }{2014}]{cho2014properties}
Kyunghyun Cho, Bart Van~Merri{\"e}nboer, Dzmitry Bahdanau, and Yoshua Bengio.
\newblock On the properties of neural machine translation: Encoder-decoder
  approaches.
\newblock {\em arXiv preprint arXiv:1409.1259}, 2014.

\bibitem[\protect\citeauthoryear{Han and Sun}{2016}]{han:2016}
Xianpei Han and Le~Sun.
\newblock Global distant supervision for relation extraction.
\newblock In {\em Thirtieth AAAI Conference on Artificial Intelligence}, 2016.

\bibitem[\protect\citeauthoryear{Hoffmann \bgroup \em et al.\egroup
  }{2011}]{hoffmann:2011}
Raphael Hoffmann, Congle Zhang, Xiao Ling, Luke Zettlemoyer, and Daniel~S Weld.
\newblock Knowledge-based weak supervision for information extraction of
  overlapping relations.
\newblock In {\em Proceedings of the 49th Annual Meeting of the Association for
  Computational Linguistics: Human Language Technologies-Volume 1}, pages
  541--550. Association for Computational Linguistics, 2011.

\bibitem[\protect\citeauthoryear{Lin \bgroup \em et al.\egroup
  }{2016}]{nre-lin}
Yankai Lin, Shiqi Shen, Zhiyuan Liu, Huanbo Luan, and Maosong Sun.
\newblock Neural relation extraction with selective attention over instances.
\newblock In {\em Proceedings of the 54th Annual Meeting of the Association for
  Computational Linguistics (Volume 1: Long Papers)}, pages 2124--2133, Berlin,
  Germany, August 2016. Association for Computational Linguistics.

\bibitem[\protect\citeauthoryear{Mintz \bgroup \em et al.\egroup
  }{2009}]{mintz:2009}
Mike Mintz, Steven Bills, Rion Snow, and Dan Jurafsky.
\newblock Distant supervision for relation extraction without labeled data.
\newblock In {\em Proceedings of the Joint Conference of the 47th Annual
  Meeting of the ACL and the 4th International Joint Conference on Natural
  Language Processing of the AFNLP: Volume 2-Volume 2}, pages 1003--1011.
  Association for Computational Linguistics, 2009.

\bibitem[\protect\citeauthoryear{Nagarajan \bgroup \em et al.\egroup
  }{2017}]{candis}
Tushar Nagarajan, Sharmistha, and Partha Talukdar.
\newblock {CANDiS}: Coupled and attention-driven neural distant supervision.
\newblock {\em arXiv preprint arXiv:1710.09942}, 10 2017.

\bibitem[\protect\citeauthoryear{Pedregosa \bgroup \em et al.\egroup
  }{2011}]{scikit-learn}
F.~Pedregosa, G.~Varoquaux, A.~Gramfort, V.~Michel, B.~Thirion, O.~Grisel,
  M.~Blondel, P.~Prettenhofer, R.~Weiss, V.~Dubourg, J.~Vanderplas, A.~Passos,
  D.~Cournapeau, M.~Brucher, M.~Perrot, and E.~Duchesnay.
\newblock Scikit-learn: Machine learning in {P}ython.
\newblock {\em Journal of Machine Learning Research}, 12:2825--2830, 2011.

\bibitem[\protect\citeauthoryear{Riedel \bgroup \em et al.\egroup
  }{2010}]{riedel:2010}
Sebastian Riedel, Limin Yao, and Andrew McCallum.
\newblock Modeling relations and their mentions without labeled text.
\newblock In {\em Machine Learning and Knowledge Discovery in Databases}, pages
  148--163. Springer, 2010.

\bibitem[\protect\citeauthoryear{Ritter \bgroup \em et al.\egroup
  }{2013}]{ritter:2013}
Alan Ritter, Luke Zettlemoyer, Oren Etzioni, et~al.
\newblock Modeling missing data in distant supervision for information
  extraction.
\newblock {\em Transactions of the Association for Computational Linguistics},
  1:367--378, 2013.

\bibitem[\protect\citeauthoryear{Shaohua~Sun and Orr}{2013}]{g:dataset}
Rahul~Gupta Shaohua~Sun, Ni~Lao and Dave Orr.
\newblock {50,000 Lessons on How to Read: a Relation Extraction Corpus}.
\newblock
  \url{https://research.googleblog.com/2013/04/50000-lessons-on-how-to-read-relation.html},
  2013.
\newblock [Online; accessed 15-NOV-2017].

\bibitem[\protect\citeauthoryear{Shen and Huang}{2016}]{byshen:2016}
Yatian Shen and Xuanjing Huang.
\newblock Attention based convolutional neural network for semantic relation
  extraction.
\newblock In {\em COLING}, 2016.

\bibitem[\protect\citeauthoryear{Srivastava \bgroup \em et al.\egroup
  }{2014}]{srivastava2014dropout}
Nitish Srivastava, Geoffrey~E Hinton, Alex Krizhevsky, Ilya Sutskever, and
  Ruslan Salakhutdinov.
\newblock Dropout: a simple way to prevent neural networks from overfitting.
\newblock {\em Journal of Machine Learning Research}, 15(1):1929--1958, 2014.

\bibitem[\protect\citeauthoryear{Surdeanu \bgroup \em et al.\egroup
  }{2012}]{surdeanu:2012}
Mihai Surdeanu, Julie Tibshirani, Ramesh Nallapati, and Christopher~D Manning.
\newblock Multi-instance multi-label learning for relation extraction.
\newblock In {\em Proceedings of the 2012 Joint Conference on Empirical Methods
  in Natural Language Processing and Computational Natural Language Learning},
  pages 455--465. Association for Computational Linguistics, 2012.

\bibitem[\protect\citeauthoryear{Weston \bgroup \em et al.\egroup
  }{2013}]{weston:2013}
Jason Weston, Antoine Bordes, Oksana Yakhnenko, and Nicolas Usunier.
\newblock Connecting language and knowledge bases with embedding models for
  relation extraction.
\newblock {\em arXiv preprint arXiv:1307.7973}, 2013.

\bibitem[\protect\citeauthoryear{Xu \bgroup \em et al.\egroup
  }{2015}]{xu+al-2015-icml}
Kelvin Xu, Jimmy Ba, Ryan Kiros, Kyunghyun Cho, Aaron Courville, Ruslan
  Salakhutdinov, Richard Zemel, and Yoshua Bengio.
\newblock Show, attend and tell: Neural image caption generation with visual
  attention.
\newblock pages 2048--2057, 2015.

\bibitem[\protect\citeauthoryear{Ye \bgroup \em et al.\egroup
  }{2016}]{DBLP:journals/corr/YeCL16}
Hai Ye, Wenhan Chao, and Zhunchen Luo.
\newblock Jointly extracting relations with class ties via effective deep
  ranking.
\newblock {\em CoRR}, abs/1612.07602, 2016.

\bibitem[\protect\citeauthoryear{Zeng \bgroup \em et al.\egroup
  }{2014}]{zeng:2014}
Daojian Zeng, Kang Liu, Siwei Lai, Guangyou Zhou, Jun Zhao, et~al.
\newblock Relation classification via convolutional deep neural network.
\newblock In {\em COLING}, pages 2335--2344, 2014.

\bibitem[\protect\citeauthoryear{Zeng \bgroup \em et al.\egroup
  }{2015}]{zeng:2015}
Daojian Zeng, Kang Liu, Yubo Chen, and Jun Zhao.
\newblock Distant supervision for relation extraction via piecewise
  convolutional neural networks.
\newblock In {\em EMNLP}, 2015.

\bibitem[\protect\citeauthoryear{Zheng \bgroup \em et al.\egroup
  }{2016}]{zheng:2016}
Hao Zheng, Zhoujun Li, Senzhang Wang, Zhao Yan, and Jianshe Zhou.
\newblock Aggregating inter-sentence information to enhance relation
  extraction.
\newblock In {\em Thirtieth AAAI Conference on Artificial Intelligence}, 2016.

\end{thebibliography}
